\documentclass{article}

\usepackage{arxiv}

\usepackage[utf8]{inputenc} 
\usepackage[T1]{fontenc}    
\usepackage{hyperref}       
\usepackage{url}            
\usepackage{booktabs}       
\usepackage{amsfonts}       
\usepackage{nicefrac}       
\usepackage{microtype}      
\usepackage{lipsum}		
\usepackage{graphicx}
\usepackage{natbib}
\usepackage{doi}
\usepackage{amsmath}
\newtheorem{theorem}{Theorem}
\newtheorem{lemma}[theorem]{Lemma}

\title{A General Method for Proving Networks Universal Approximation Property}

\author{ Wei Wang
}

\begin{document}
\maketitle

\begin{abstract}
Deep learning architectures are highly diverse. To prove their universal approximation properties, existing works typically rely on model-specific proofs. Generally, they construct a dedicated mathematical formulation for each architecture (e.g., fully connected networks, CNNs, or Transformers) and then prove their universal approximability. However, this approach suffers from two major limitations: first, every newly proposed architecture often requires a completely new proof from scratch; second, these proofs are largely isolated from one another, lacking a common analytical foundation. This not only incurs significant redundancy but also hinders unified theoretical understanding across different network families. To address these issues, this paper proposes a general and modular framework for proving universal approximation. We define a basic building block (comprising one or multiple layers) that possesses the universal approximation property as a Universal Approximation Module (UAM). Under this condition, we show that any deep network composed of such modules inherently retains the universal approximation property. Moreover, the overall approximation process can be interpreted as a progressive refinement across modules. This perspective not only unifies the analysis of diverse architectures but also enables a step-by-step understanding of how expressive power evolves through the network.
\end{abstract}


\section{Introduction}

In recent years, deep learning has achieved remarkable success in fields such as computer vision and natural language processing. However, this success has largely relied on empirical validation and engineering optimization rather than solid theoretical foundations. The fundamental reason for this gap lies in the high complexity of deep neural network architectures, nonlinearity, high dimensionality, and hierarchical coupling make it extremely challenging to rigorously characterize their behavior using formal mathematics.

From a theoretical perspective, the effectiveness of deep learning is widely attributed to its powerful function approximation capability. To further enhance model expressivity and generalization performance, researchers have proposed a variety of network architectures, including fully connected neural networks (FCNNs) \cite{Gardner1998ArtificialNN,Murtagh1991MultilayerPF}, convolutional neural networks (CNNs) \cite{Shelhamer2014FullyCN, He2015DeepRL}, and more recently, the widely adopted Transformer \cite{Vaswani2017AttentionIA,Dosovitskiy2020AnII} architecture. Although these models differ significantly in design motivation or application scenarios, they share a common goal: to effectively approximate complex functions through specific parameterized structures.

Consequently, a central direction in the theoretical study of deep learning is the systematic analysis of the approximation capacity of different network architectures. Existing works \cite{Cybenko1989ApproximationBS, Hornik1989MultilayerFN,Hornik1991ApproximationCO} have established universal approximation theorem (UAT). Recently, model-specific UATs have been developed for architectures of CNNs \cite{Zhou2018UniversalityOD,Oono2019ApproximationAN} and Transformers \cite{Yun2019AreTU}, in order to account for their distinct structural properties. A typical research paradigm begins by analyzing single-hidden-layer architectures, proving that they can approximate any continuous function under certain conditions, and then gradually extends these results to multi-layer or deep settings. Such approaches usually treat the entire network as a single mathematical mapping and construct a formal approximation framework to demonstrate its universal approximation property.

While these studies have made significant contributions to understanding the theoretical principles of deep learning, two key limitations remain. First, existing theories are highly architecture-specific and lack generality. When a new architecture emerges, researchers must rebuild the mathematical model from scratch and repeat complex proofs, making it difficult to develop a unified analytical framework. Second, these methods, although capable of proving approximation power by abstracting the entire network as a "black-box" composite function, fail to reveal the internal learning dynamics or information evolution mechanisms across layers or modules. Precisely, the insights are needed to understand model behavior and guide architectural design and optimization.

To address these challenges, this paper proposes a more general theoretical framework. Our core idea is to move beyond the specifics of individual network architectures and instead take universal approximation property (UAP) as a foundational assumption. We abstract any substructure within a network, such as a single layer or a multi-layer block, that possesses this property into a Universal Approximation Module (UAM). Building upon this abstraction, we develop a modular theory of approximation that not only provides a unified proof of the overall universal approximation capability for diverse network architectures but also enables a step-by-step analysis of how information and representations evolve throughout the approximation process.

The contribution of this paper as following:

\begin{itemize}
\item This paper builds up a general approch for proving the universal approximation property of a neural network. It eliminates dependence on specific architectural details. As long as a component is proven to be universally approximating (regardless of its internal structure), it can be seamlessly integrated into our framework for compositional analysis. 

\item This paper give the proof from the modular perspective. It allows us to trace the internal approximation pathway of the network, offering a theoretical lens for understanding the inner workings of deep models. This not only shedding light on how representations evolve across modules but also provides an extensible analytical paradigm for the design and theoretical validation of future neural network models.
\end{itemize}

\section{Theoretical Analysis of Progressive Approximation in Module Networks}

\subsection{Problem Setup}
Let the input space $\mathcal{X} \subseteq \mathbb{R}^m$ be a compact set, and the target function $f: \mathcal{X} \to \mathbb{R}^m$ be continuous. Consider a neural network architecture composed of $N$ cascaded learnable modules:
$$
\mathbf{x}_i = H_i(\mathbf{x}_{i-1}), \quad i = 1, 2, \dots, N,
$$
where the initial input is $\mathbf{x}_0 \in \mathcal{X}$, each mapping $H_i: \mathbb{R}^m \to \mathbb{R}^m$ belongs to some function class $\mathcal{H}_i$, and all intermediate representations satisfy $\mathbf{x}_i \in \mathbb{R}^m$.

The overall prediction function is defined as the composition:
$$
\hat{f} := H_N \circ H_{N-1} \circ \cdots \circ H_1,
$$
so that $\hat{f}(\mathbf{x}_0) = \mathbf{x}_N$. The central question of this paper is how such a cascaded structure can approximate the continuous function $f$ to arbitrary accuracy? More precisely, we aim to establish an approximation property where, for any $\varepsilon > 0$, there exists a positive integer $N$ and a sequence of functions $\{H_i^*\}_{i=1}^N$ (with $H_i^* \in \mathcal{H}_i$) such that
$$
\sup_{\mathbf{x}_0 \in \mathcal{X}} \| \hat{f}(\mathbf{x}_0) - f(\mathbf{x}_0) \| < \varepsilon.
$$

\subsection{Key Assumptions and Preliminaries}

We begin with a standard expressiveness assumption on each module.

\textbf{Assumption 1 (Universal Approximation Property, UAP).}
Each continuous function class $\mathcal{H}_i$ is dense in $C(\mathcal{K}_i, \mathbb{R}^m)$ for any non-empty compact set $\mathcal{K}_i \subset \mathbb{R}^m$. That is, for any continuous $g: \mathcal{K}_i \to \mathbb{R}^m$ and $\varepsilon >0$, there exists $H_i^* \in \mathcal{H}_i$ such that
$$
\sup_{\mathbf{x}_{i-1} \in \mathcal{K}_i} \|H_i^*(\mathbf{x}_{i-1}) - g(\mathbf{x}_{i-1})\| < \varepsilon.
$$

This condition ensures that each module has sufficient expressive power and serves as the foundation for constructing the approximation path. Notably, many common model classes satisfy this property, such as sufficiently wide feedforward neural networks \cite{Cybenko1989ApproximationBS, Hornik1989MultilayerFN,Hornik1991ApproximationCO}.

\textbf{Property (Compactness Preservation).} If a function $H_i: \mathbb{R}^m \to \mathbb{R}^m$ is continuous and $\mathcal{K}_{i-1} \subset \mathbb{R}^m$ is compact, then the image $H_i(\mathcal{K}_{i-1})$ is also compact. This property follows directly from the fact that continuous maps preserve compactness. Therefore, it is not an additional assumption but an inherent property of continuous functions.

In this work, we repeatedly use this property to ensure the compactness of input and output sets at each module, thereby allowing the application of universal approximation theorems at every step.

\textbf{Condition (Target Recoverability, TR).} At module $i$, we define the encoding map at step $i$ as:
$$
\phi_i := H_{i} \circ \cdots \circ H_1, \quad \text{for } i=1\cdots n.
$$
For the final output $\mathbf{x}_N = \hat{f}(\mathbf{x}_0)$ to approximate $f(\mathbf{x}_0)$, it is essential that the information about $f(\mathbf{x}_0)$ is presered throughout the intermediate representations.

To this end, we propose the following necessary condition: for any $\mathbf{x}_0^1, \mathbf{x}_0^2 \in \mathcal{X}$ and any $i=1 \cdots N$, if $\phi_i(\mathbf{x}_0^1) = \phi_i(\mathbf{x}_0^2)$, then $f(\mathbf{x}_0^1) = f(\mathbf{x}_0^2)$ must hold. In other words, the equivalence relation induced by $\phi_i$ must not be coarser than that induced by $f$. Equivalently, there exists a continuous function $h_i: \mathbb{R}^m \to \mathbb{R}^m$ such that
$$
f = h_i \circ \phi_i \quad \text{on } \mathcal{X}.
$$

We refer to this property as Target Recoverability (TR), which is a prerequisite for layer-wise approximation. If a representation $\mathbf{x}_{i}$ fails to distinguish inputs with different target values, subsequent modules cannot recover this information, leading to approximation failure.

\subsection{Main Result}
The main result establishes that deep modular networks can progressively refine their approximation to the target function, provided each module is sufficiently expressive (UAP) and information about the target is preserved throughout the cascade (TR). Unlike classical universal approximation theorems that treat a network as a whole (black box), our result highlights the feasibility of layer-wise approximation and information-preserving in a constructive manner. Based on the above conditions, this paper obtain the following main theorem:

\begin{theorem}[Modular Approximation]  
\label{th:modular_approximation}
Let $\mathcal{X} \subseteq \mathbb{R}^m$ be compact, $f: \mathcal{X} \to \mathbb{R}^m$ be continuous, and each function class $\mathcal{H}_i$ satisfy the Universal Approximation Property (UAP). Then, for any $\varepsilon>0$, there exists a positive integer $N$ and a sequence of functions $\{H_i^*\}_{i=1}^N$ (with $H_i^* \in \mathcal{H}_i$), such that the overall mapping defined by
$$
\hat{f} = H_N^* \circ \cdots \circ H_1^*.
$$
The approximation error decreases strictly across layers:
$$
\sup_{\mathbf{x}_0 \in \mathcal{X}} \| \mathbf{x}_i - f(\mathbf{x}_0) \| < \varepsilon_i, \quad \text{with } \varepsilon_1 > \varepsilon_2 > \cdots > \varepsilon_N \text{ and } \varepsilon_N < \varepsilon,
$$
and Target Recoverability holds at every layer.
\end{theorem}

\noindent
\textit{Note:} Unlike classical universal approximation results, this theorem highlights the feasibility of progressive, stage-wise approximation under information preservation. The proof is constructive and reveals how deep cascaded modules can iteratively refine the estimate of $f$.

\subsection{Proof Sketch: Constructing a Progressive Approximation Path}

The main idea is to construct the network module by module, maintaining two invariants at each step:
\begin{itemize}
    \item The output $\mathbf{x}_i = \phi_i(\mathbf{x}_0)$ approximates $f(\mathbf{x}_0)$ within a decreasing error bound $\varepsilon_i$;
    \item The output of $\phi_i$ satisfies Target Recoverability (TR), i.e., $\phi_i \in \mathcal{T}_f = \{ \phi \in C(\mathcal{X}, \mathbb{R}^m) \mid \forall \mathbf{x}^1, \mathbf{x}^2 \in \mathcal{X}, \phi(\mathbf{x}^1)=\phi(\mathbf{x}^2) \Rightarrow f(\mathbf{x}^1)=f(\mathbf{x}^2) \}$
\end{itemize}


Let $\varepsilon_1 = 1$. Since $\mathcal{H}_1$ is dense in $C(\mathcal{X}, \mathbb{R}^m)$ and $\mathcal{T}_f$ is dense, their intersection is nonempty in any neighborhood of $f$. Thus, there exists $H_1^* \in \mathcal{H}_1 \cap \mathcal{T}_f$ such that
$$
\sup_{\mathbf{x}_0 \in \mathcal{X}} \| H_1^*(\mathbf{x}_0) - f(\mathbf{x}_0) \| < \varepsilon_1.
$$
Set $\phi_1 = H_1^*$ and $\mathcal{K}_1 = \phi_1(\mathcal{X})$, which is compact.

Inductively, suppose after $i$ layers we have $\phi_i = H_i^* \circ \cdots \circ H_1^*$ such that:
$$
\|\phi_i(\mathbf{x}_0) - f(\mathbf{x}_0)\| < \varepsilon_i \quad \text{and} \quad \phi_i \in \mathcal{T}_f.
$$

By TR, there exists a continuous $h_i: \mathcal{K}_i \to \mathbb{R}^m$ (with $\mathcal{K}_i = \phi_i(\mathcal{X})$ compact) such that $f = h_i \circ \phi_i$. The UAP of $\mathcal{H}_{i+1}$ then guarantees the existence of $H_{i+1}^* \in \mathcal{H}_{i+1} \cap \mathcal{T}_{h_i}$ satisfying
$$
\sup_{\mathbf{z} \in \mathcal{K}_i} \|H_{i+1}^*(\mathbf{z}) - h_i(\mathbf{z})\| < \varepsilon_{i+1},
$$
which yields a refined estimate $\phi_{i+1} = H_{i+1}^* \circ \phi_i$ with error $\varepsilon_{i+1} < \varepsilon_i$ and preserved TR.

Repeating this process with $\varepsilon_i = \varepsilon_1 / 2^{i-1}$, we achieve error below any $\varepsilon > 0$ in finite depth $N$. The full construction—including precise error bounds and preservation of compactness (More details in Appendix ).

\subsection{Discussion}
Our result establishes that deep modular networks can approximate any continuous function progressively, provided two key ingredients:
\begin{itemize}
    \item \textbf{Expressiveness} (UAP): Each module can approximate arbitrary continuous mappings on compact domains;
    \item \textbf{Information Preservation} (TR): No layer collapses inputs that differ in their target values.
\end{itemize}

This contrasts with classical universal approximation theorems, which assert existence without specifying how the approximation unfolds across layers. Here, we reveal a feasible approximation path where error decreases monotonically and information about the target is preserved throughout.

For convenience, the proof above adopts a geometrically decaying error sequence (i.e., \(\varepsilon_i = \varepsilon_{i-1}/2\)). In practice, however, factors such as optimization methods, module capacity, and data distribution can influence the actual approximation path, which as linear or other forms of error decay. Nevertheless, our analysis provides a theoretical foundation for understanding the module-wise refinement mechanism in deep cascaded models.

Thus, the combination of UAP and TR offers a sufficient condition for both expressiveness and structured learnability in deep modular architectures.

\section{Probability Approximation: Modular Learning in Classification Tasks}

Let the input space $\mathcal{X} \subseteq \mathbb{R}^d$ be compact, and let the target posterior distribution $\mathbf{p}: \mathcal{X} \to \Delta^{n-1}$ be a continuous function such that $p_k(\mathbf{x}) > 0$ for all classes $k = 1,\dots,n$ and all $\mathbf{x} \in \mathcal{X}$. Consider a modular network of the form
$$
\hat{\mathbf{p}}(\mathbf{x}) = \mathrm{Softmax}\big(W_{N+1} \circ H_N \circ \cdots \circ H_1(\mathbf{x})\big),
$$
where each $H_i$ belongs to a universal approximator function class $\mathcal{H}_i$ (we have already established that the composition $H_N \circ \cdots \circ H_1$ is a universal approximator). Moreover, the entire mapping $W_{N+1} \circ H_N \circ \cdots \circ H_1(\mathbf{x})$ constitutes a universal approximator. (In general, universal approximator classes $\mathcal{H}_i$ are equipped with linear or affine output layers, so the final linear map $W_{N+1}$ can be absorbed into $H_N$ as part of its definition.)

Then, for any $\varepsilon > 0$, there exist a positive integer $N$, a sequence of functions $\{H_i\}_{i=1}^N$ with $H_i \in \mathcal{H}_i$, and a weight matrix $W_{N+1}^ \in \mathbb{R}^{n \times m}$ such that
$$
\sup_{\mathbf{x} \in \mathcal{X}} \|\hat{\mathbf{p}}(\mathbf{x}) - \mathbf{p}(\mathbf{x})\| < \varepsilon.
$$

\noindent \textbf{Proof.}

We define the logits corresponding to $\mathbf{p}$ as
$$
z_k(\mathbf{x}) := \log p_k(\mathbf{x}) - \frac{1}{n} \sum_{j=1}^n \log p_j(\mathbf{x}), \quad k = 1, \dots, n.
$$
By the definition of the Softmax function, we have
$$
\mathbf{p}(\mathbf{x}) = \mathrm{Softmax}(\mathbf{z}(\mathbf{x})) := \left( \frac{e^{z_1(\mathbf{x})}}{\sum_j e^{z_j(\mathbf{x})}}, \dots, \frac{e^{z_n(\mathbf{x})}}{\sum_j e^{z_j(\mathbf{x})}} \right).
$$

Define the composite mapping

$$
f(\mathbf{x}) := W_{N+1} \circ H_N \circ \cdots \circ H_1(\mathbf{x}) \in \mathbb{R}^n.
$$

The function class

$$
\mathcal{F} = \left \{ f = W_{N+1} \circ H_N \circ \cdots \circ H_1 \,\cdots \, N \in \mathbb{N},\, H_i \in \mathcal{H}_i,\, W_{N+1} \in \mathbb{R}^{n \times m} \right \}
$$

is a universal approximator. This means that for any continuous function $\mathbf{g}: \mathcal{X} \to \mathbb{R}^n$ and any $\delta > 0$, there exists some $f \in \mathcal{F}$ such that
$$
\sup_{\mathbf{x} \in \mathcal{X}} \| f(\mathbf{x}) - \mathbf{g}(\mathbf{x}) \| < \delta.
$$
In particular, setting $\mathbf{g} = \mathbf{z}$, for any $\delta > 0$ there exists a function $\hat{f} = W_{N+1} \circ H_N \circ \cdots \circ H_1$ satisfying
$$
\sup_{\mathbf{x} \in \mathcal{X}} \| \hat{f}(\mathbf{x}) - \mathbf{z}(\mathbf{x}) \| < \delta.
$$

Since the Softmax function is Lipschitz continuous on any bounded subset of its domain, and since $\mathcal{X}$ is compact and $\mathbf{z}$ is continuous, the image $\mathbf{z}(\mathcal{X})$ is bounded. By choosing $\hat{f}$ sufficiently close to $\mathbf{z}$, we may assume $\hat{f}(\mathcal{X})$ lies within a common bounded set containing $\mathbf{z}(\mathcal{X})$. Let $L > 0$ denote a Lipschitz constant of Softmax on this set.

Now define $\hat{\mathbf{p}}(\mathbf{x}) = \mathrm{Softmax}(\hat{f}(\mathbf{x}))$. Then for all $\mathbf{x} \in \mathcal{X}$,
$$
\| \hat{\mathbf{p}}(\mathbf{x}) - \mathbf{p}(\mathbf{x}) \|
= \| \mathrm{Softmax}(\hat{f}(\mathbf{x})) - \mathrm{Softmax}(\mathbf{z}(\mathbf{x})) \|
\leq L \| \hat{f}(\mathbf{x}) - \mathbf{z}(\mathbf{x}) \| .
$$
Taking the supremum over $\mathcal{X}$ yields
$$
\sup_{\mathbf{x} \in \mathcal{X}} \| \hat{\mathbf{p}}(\mathbf{x}) - \mathbf{p}(\mathbf{x}) \|
\leq L \cdot \sup_{\mathbf{x} \in \mathcal{X}} \| \hat{f}(\mathbf{x}) - \mathbf{z}(\mathbf{x}) \|
< L \delta.
$$
Given any $\varepsilon > 0$, choose $\delta = \varepsilon / L$. Then there exists a network architecture (i.e., suitable $N$, $\{H_i\}_{i=1}^N$, and $W_{N+1}$) such that
$$
\sup_{\mathbf{x} \in \mathcal{X}} \| \hat{\mathbf{p}}(\mathbf{x}) - \mathbf{p}(\mathbf{x}) \| < \varepsilon,
$$
which completes the proof.

\section{Conclusion}
This paper proves that if the basic blocks (composed of one or multiple layers) of a neural network owns the universal approximation property, then a deep network formed by cascading multiple such modules also retains universal approximability. Moreover, the overall approximation process can be understood as a progressive, module-by-module refinement, in which the network gradually enhances its expressivity and approximation accuracy for the target function through the composition of these modules.

\bibliographystyle{unsrtnat}
\bibliography{references} 
\appendix

\section{Constructive Proof}

We prove the theorem \ref{th:modular_approximation} by explicitly constructing a sequence of modules $\{H_i^*\}_{i=1}^N$ such that the composition $\hat{f} = H_N^* \circ \cdots \circ H_1^*$ approximates $f$ to arbitrary precision, while preserving Target Recoverability (TR) at every layer. The construction proceeds by induction.

\paragraph{Preliminaries.}
We assume all middle modules (except for the first module) map $\mathbb{R}^m \to \mathbb{R}^m$, consistent with the problem setup. Key ingredients:
\begin{itemize}
    \item Each $\mathcal{H}_i$ consists of continuous functions and satisfies the UAP on any compact domain $\mathcal{K} \subset \mathbb{R}^m$;
    \item Since $\mathcal{X}$ is compact and all $H_i^*$ are continuous, each intermediate representation space $\mathcal{K}_i = \phi_i(\mathcal{X})$ is compact;
    \item Let $\mathcal{T}_f = \{ \phi \in C(\mathcal{X}, \mathbb{R}^m) \mid \forall \mathbf{x}^1, \mathbf{x}^2 \in \mathcal{X}, \phi(\mathbf{x}^1)=\phi(\mathbf{x}^2) \Rightarrow f(\mathbf{x}^1)=f(\mathbf{x}^2) \}$. By Lemma~\ref{TR_dense} (Appendix), $\mathcal{T}_f$ is dense in $C(\mathcal{X}, \mathbb{R}^m)$.
\end{itemize}

\paragraph{Base Case.}
Let $\varepsilon_1 = 1$. Since $\mathcal{H}_1$ is dense in $C(\mathcal{X}, \mathbb{R}^m)$ and $\mathcal{T}_f$ is dense, their intersection is nonempty in any neighborhood of $f$. Thus, there exists $H_1^* \in \mathcal{H}_1 \cap \mathcal{T}_f$ such that
$$
\sup_{\mathbf{x}_0 \in \mathcal{X}} \| H_1^*(\mathbf{x}_0) - f(\mathbf{x}_0) \| < \varepsilon_1.
$$
Set $\phi_1 = H_1^*$ and $\mathcal{K}_1 = \phi_1(\mathcal{X})$, which is compact.

\paragraph{Inductive Step.}
Assume for some $i \geq 1$ that $\phi_i = H_i^* \circ \cdots \circ H_1^*$ satisfies:
\begin{itemize}
    \item $\sup_{\mathbf{x}_0 \in \mathcal{X}} \| \phi_i(\mathbf{x}_0) - f(\mathbf{x}_0) \| < \varepsilon_i$;
    \item $\phi_i \in \mathcal{T}_f$;
    \item $\mathcal{K}_i = \phi_i(\mathcal{X})$ is compact.
\end{itemize}
By the inductive hypothesis, $\phi_i \in \mathcal{T}_f$, so there exists a unique continuous map $h_i: \mathcal{K}_i \to \mathbb{R}^m$ such that $f = h_i \circ \phi_i$. Moreover, since $\phi_i$ and $f$ have identical fibers, a function $H \in C(\mathcal{K}_i, \mathbb{R}^m)$ satisfies $H \circ \phi_i \in \mathcal{T}_f$ if and only if $H \in \mathcal{T}_{h_i}$, where  
$$
\mathcal{T}_{h_i} := \{ H \in C(\mathcal{K}_i, \mathbb{R}^m) \mid \forall \mathbf{z}^1, \mathbf{z}^2 \in  \mathcal{K}_i, H(\mathbf{z}^1)=H(\mathbf{z}^2) \Rightarrow h_i(\mathbf{z}^1)=h_i(\mathbf{z}^2) \}.$$
Since $\mathcal{K}_i$ is compact and $h_i$ is continuous, Lemma~\ref{TR_dense} applies to $(\mathcal{K}_i, h_i)$, implying that $\mathcal{T}_{h_i}$ is dense and contains an open neighborhood of $h_i$ in $C(\mathcal{K}_i, \mathbb{R}^m)$. Set $\varepsilon_{i+1} = \varepsilon_i / 2$. Since $\mathcal{H}_{i+1}$ is dense in $C(\mathcal{K}_i, \mathbb{R}^m)$ by the UAP assumption, there exists $H_{i+1}^* \in \mathcal{H}_{i+1} \cap \mathcal{T}_{h_i}$ such that
$$
\sup_{\mathbf{z} \in \mathcal{K}_i} \| H_{i+1}^*(\mathbf{z}) - h_i(\mathbf{z}) \| < \varepsilon_{i+1}.
$$
Define $\phi_{i+1} = H_{i+1}^* \circ \phi_i$. Then we have $\| \phi_{i+1}(\mathbf{x}_0) - f(\mathbf{x}_0) \| < \varepsilon_{i+1}$, $\phi_{i+1} \in \mathcal{T}_f$, and $\mathcal{K}_{i+1} = \phi_{i+1}(\mathcal{X})$ is compact.  

\textbf{Global Approximation}

Define the error sequence $\varepsilon_i = \varepsilon_1 / 2^{i-1}$, which converges to 0. For any $\varepsilon > 0$, choose
$$
N = \left\lceil \log_2\left( \frac{\varepsilon_1}{\varepsilon} \right) \right\rceil + 1,
$$
so that $\varepsilon_N < \varepsilon$. Then the final mapping $\hat{f} = \phi_N$ satisfies
$$
\sup_{\mathbf{x}_0 \in \mathcal{X}} \| \hat{f}(\mathbf{x}_0) - f(\mathbf{x}_0) \| < \varepsilon_N < \varepsilon.
$$

By construction, TR holds at every layer $i = 1, \dots, N$, and each $\mathcal{K}_i$ is compact. Therefore, the sequence $\{H_i^*\}_{i=1}^N$ achieves arbitrary approximation accuracy while preserving target information throughout the network. This completes the proof of both the theorem and the corollary.



\section{Density of Target-Recoverable Functions}
Let $\mathcal{X} \subset \mathbb{R}^d$ be a nonempty compact metric space equipped with the Euclidean distance, and let $f \colon \mathcal{X} \to \mathbb{R}^m$ be a continuous function.  
We define the set of $f$-distinguished point pairs as
$$
\mathcal{P}_f := \left\{ (\mathbf{x}^1, \mathbf{x}^2) \in \mathcal{X} \times \mathcal{X} \,\middle|\, \mathbf{x}^1 \ne \mathbf{x}^2 \text{ and } f(\mathbf{x}^1) \ne f(\mathbf{x}^2) \right\};
$$
A continuous function $G \colon \mathcal{X} \to \mathbb{R}^m$ is said to satisfy Target Recoverability (TR) with respect to $f$ if it preserves all distinctions made by $f$; that is, if
$$
\forall (\mathbf{x}^1, \mathbf{x}^2) \in \mathcal{P}_f, \quad  G(\mathbf{x}^1) \neq G(\mathbf{x}^2).
$$
The collection of all such functions is denoted by
$$
\mathcal{T}_f := \left\{ G \in C(\mathcal{X}, \mathbb{R}^m) \,\middle|\, \forall\, (\mathbf{x}^1, \mathbf{x}^2) \in \mathcal{P}_f,\ G(\mathbf{x}^1) \ne G(\mathbf{x}^2) \right\}.
$$

The space $C(\mathcal{X}, \mathbb{R}^m)$ is endowed with the uniform norm
$$
\|G\|_\infty = \sup_{\mathbf{x} \in \mathcal{X}} \|G(\mathbf{x})\|,
$$
under which it becomes a complete metric space.

\begin{lemma}
The set \(\mathcal{T}_f\) is dense in \(C(\mathcal{X}, \mathbb{R}^m)\). Consequently:
\begin{itemize}
    \item For every \(\varepsilon > 0\), the open ball \(B_\varepsilon(f)\) satisfies \(B_\varepsilon(f) \cap \mathcal{T}_f \ne \emptyset\);
    \item The complement \(C(\mathcal{X}, \mathbb{R}^m) \setminus \mathcal{T}_f\) is a meager set (i.e., a countable union of nowhere dense sets).
\end{itemize}
\label{TR_dense}
\end{lemma}

\textbf{Proof:}

\textbf{1. Construction of “bad” function sets.} 

For each $n \in \mathbb{N}$, define the compact set of well-separated point pairs at scale $1/n$ by
$$
\mathcal{D}_n := \left\{ (\mathbf{x}^1, \mathbf{x}^2) \in \mathcal{X} \times \mathcal{X} \,\middle|\, \|\mathbf{x}^1 - \mathbf{x}^2\| \ge \frac{1}{n} \right\}.
$$
Since \(\mathcal{X}\) is compact, \(\mathcal{D}_n\) is a closed subset of \(\mathcal{X} \times \mathcal{X}\), hence compact.

Define the “bad” function set
$$
\mathcal{F}_n := \left\{ G \in C(\mathcal{X}, \mathbb{R}^m) \,\middle|\, \exists\, (\mathbf{x}^1, \mathbf{x}^2) \in \mathcal{D}_n \text{ such that } G(\mathbf{x}^1) = G(\mathbf{x}^2) \right\}.
$$

\noindent\textit{Note.} The sets $\mathcal{F}_n$ capture all continuous functions that fail to be injective on at least one pair of points separated by distance at least $1/n$. The positive separation threshold ensures that each $F_n$ is closed in the uniform topology which is a key property for the subsequent Baire category argument.


\textbf{2. Each \(\mathcal{F}_n\) is a closed, nowhere dense set.}

We now show that each $F_n$ is closed and nowhere dense in $C(\mathcal{X}, \mathbb{R}^m)$.

\emph{\textbf{Closedness}.}  
Let $\{G_k\}_{k=1}^\infty \subset \mathcal{F}_n$ be a sequence converging uniformly to some $G \in C(\mathcal{X}, \mathbb{R}^m)$.  
For each $k$, there exists $(\mathbf{x}_k^1, \mathbf{x}_k^2) \in \mathcal{D}_n$ such that $G_k(\mathbf{x}^1_k) = G_k(\mathbf{x}^2_k)$.  
Since $\mathcal{D}_n$ is compact, we may extract a subsequence (still denoted $(\mathbf{x}^1_k, \mathbf{x}^2_k)$) converging to some $(\mathbf{x}^1, \mathbf{x}^2) \in \mathcal{D}_n$.  
By uniform convergence and continuity of $G$, we have
$$
G(\mathbf{x}^1) = \lim_{k \to \infty} G_k(\mathbf{x}^1_k) = \lim_{k \to \infty} G_k(\mathbf{x}^2_k) = G(\mathbf{x}^2).
$$
Thus $G(\mathbf{x}^1) = G(\mathbf{x}^2)$ with $(\mathbf{x}^1, \mathbf{x}^2) \in \mathcal{D}_n$, so $G \in F_n$. Hence $\mathcal{F}_n$ is closed.

\emph{\textbf{Empty interior (nowhere dense).}}  
To show that $\mathcal{F}_n$ has empty interior, fix an arbitrary $G \in \mathcal{F}_n$ and $\delta > 0$. We construct a perturbation $\tilde{G} \in B_\delta(G)$ such that $\tilde{G} \notin \mathcal{F}_n$.

Recall that $\mathcal{D}_n \subset \mathcal{X} \times \mathcal{X}$ is compact and satisfies $\|\mathbf{x}^1 - \mathbf{x}^2\| \ge 1/n > 0$ for all $(\mathbf{x}^1, \mathbf{x}^2) \in \mathcal{D}_n$.  
Define the continuous map $\psi \colon \mathcal{X} \to \mathbb{R}^m$ by
$$
\psi(\mathbf{x}) := \frac{1}{L} (\mathbf{x}, 0, \dots, 0), \quad \text{where } L := \sup_{\mathbf{x} \in \mathcal{X}} \|\mathbf{x}\| + 1.
$$
Then $\|\psi\|_\infty \le 1$, and for all $(\mathbf{x}^1, \mathbf{x}^2) \in \mathcal{D}_n$,
$$
\|\psi(\mathbf{x}^1) - \psi(\mathbf{x}^2)\| \ge \frac{1}{L} \|\mathbf{x}^1 - \mathbf{x}^2\| \ge \frac{1}{Ln} > 0.
$$

For $\lambda \in [0,1]$, define the perturbed function
$$
G_\lambda(\mathbf{x}) := G(\mathbf{x}) + \lambda \psi(\mathbf{x}), \quad \mathbf{x} \in \mathcal{X},
$$
and consider the continuous function
$$
F \colon \mathcal{D}_n \times [0,1] \to \mathbb{R}, \quad
F(\mathbf{x}^1, \mathbf{x}^2, \lambda) := \big\| G_\lambda(\mathbf{x}^1) - G_\lambda(\mathbf{x}^2) \big\|.
$$
Our goal is to find some $\lambda_0 \in (0, \delta)$ such that $F(\cdot, \lambda_0) > 0$ on all of $\mathcal{D}_n$; then $\tilde{G} := G_{\varepsilon_0} \notin \mathcal{F}_n$ and $\|\tilde{G} - G\|_\infty \le \lambda_0 < \delta$.

We now establish this by a compactness argument. Fix any $(\mathbf{x}^1, \mathbf{x}^2) \in \mathcal{D}_n$. There are two mutually exclusive cases:

\noindent\textbf{Case 1: $G(\mathbf{x}^1) = G(\mathbf{x}^2)$.}  
Then for any $\lambda > 0$,
$$
F(\mathbf{x}^1, \mathbf{x}^2, \lambda) = \lambda \|\psi(\mathbf{x}^1) - \psi(\mathbf{x}^2)\| \ge \lambda \cdot \frac{1}{Ln} > 0.
$$
Thus, for this pair, $F > 0$ holds for all $\lambda > 0$. In particular, we may choose any $\lambda_{\mathbf{x}^1,\mathbf{x}^2} > 0$ (for instance, $\lambda_{\mathbf{x}^1,\mathbf{x}^2} = 1$); the positivity is guaranteed regardless of the choice.

\noindent\textbf{Case 2: $G(\mathbf{x}^1) \ne G(\mathbf{x}^2)$.}  
Then $a := F(\mathbf{x}^1, \mathbf{x}^2, 0) = \|G(\mathbf{x}^1) - G(\mathbf{x}^2)\| > 0$.  
Since $F$ is continuous on $\mathcal{D}_n \times [0,1]$, it is continuous at $((\mathbf{x}^1, \mathbf{x}^2), 0)$. Hence, there exists a neighborhood of this point on which $F$ stays strictly positive. 
In particular, we can find $\eta > 0$ and a positive number $\lambda_{\mathbf{x}^1,\mathbf{x}^2} > 0$ (depending on the pair $(\mathbf{x}^1, \mathbf{x}^2)$) such that for all  
$$
(\mathbf{y}^1, \mathbf{y}^2) \in B_\eta(\mathbf{x}^1, \mathbf{x}^2) \cap \mathcal{D}_n \quad \text{and} \quad \lambda \in [0, \lambda_{\mathbf{x}^1,\mathbf{x}^2}),
$$  
we have
$$
|F(\mathbf{y}^1, \mathbf{y}^2, \lambda) - a| < \frac{a}{2}.
$$
Consequently,
$$
F(\mathbf{y}^1, \mathbf{y}^2, \lambda) > \frac{a}{2} > 0.
$$
Thus, on the open neighborhood $U_{\mathbf{x}^1,\mathbf{x}^2} := B_\eta(\mathbf{x}^1, \mathbf{x}^2) \cap \mathcal{D}_n$, the function $F(\cdot, \lambda)$ remains strictly positive for all $\lambda < \lambda_{\mathbf{x}^1,\mathbf{x}^2}$.

In both cases, for each $(\mathbf{x}^1, \mathbf{x}^2) \in \mathcal{D}_n$, we have constructed an open neighborhood $U_{\mathbf{x}^1,\mathbf{x}^2} \subset \mathcal{D}_n$ and a number $\lambda_{\mathbf{x}^1,\mathbf{x}^2} > 0$ such that  
$$
F(\mathbf{y}^1, \mathbf{y}^2, \lambda) > 0 \quad \text{for all } (\mathbf{y}^1, \mathbf{y}^2) \in U_{\mathbf{x}^1,\mathbf{x}^2},\ \lambda \in [0, \lambda_{\mathbf{x}^1,\mathbf{x}^2}).
$$

Since $\mathcal{D}_n$ is compact, the open cover $\{U_{\mathbf{x}^1,\mathbf{x}^2}\}_{(\mathbf{x}^1,\mathbf{x}^2) \in \mathcal{D}_n}$ admits a finite subcover:
$$
\mathcal{D}_n \subset \bigcup_{i=1}^N U_i,
$$
with corresponding positive numbers $\lambda_1, \dots, \lambda_N$. Set
$$
\lambda_0 := \min\left\{ \frac{\delta}{2},\ \lambda_1, \dots, \lambda_N \right\} > 0.
$$
Then for every $(\mathbf{x}^1, \mathbf{x}^2) \in \mathcal{D}_n$, there exists $i$ such that $(\mathbf{x}^1, \mathbf{x}^2) \in U_i$, and hence
$$
F(\mathbf{x}^1, \mathbf{x}^2, \lambda_0) > 0.
$$
That is, $G_{\lambda_0}(\mathbf{x}^1) \ne G_{\lambda_0}(\mathbf{x}^2)$ for all $(\mathbf{x}^1, \mathbf{x}^2) \in \mathcal{D}_n$, so $G_{\lambda_0} \notin \mathcal{F}_n$. Finally, since $\|\psi\|_\infty \le 1$, we have
$$
\|G_{\lambda_0} - G\|_\infty = \lambda_0 \|\psi\|_\infty \le \lambda_0 < \delta,
$$
so $G_{\lambda_0} \in B_\delta(G)$.

Because $G \in \mathcal{F}_n$ and $\delta > 0$ were arbitrary, no open ball centered at a point of $\mathcal{F}_n$ is entirely contained in $\mathcal{F}_n$. Therefore, $\mathcal{F}_n$ has empty interior. Combined with its closedness, this implies that $\mathcal{F}_n$ is nowhere dense in $C(\mathcal{X}, \mathbb{R}^m)$.

\textbf{3. The complement of \(\mathcal{T}_f\) is contained in a countable union of the sets \(\mathcal{F}_n\).}

We now establish the key inclusion
$$
C(\mathcal{X}, \mathbb{R}^m) \setminus \mathcal{T}_f \subseteq \bigcup_{n=1}^\infty \mathcal{F}_n.
$$

Indeed, let \(G \in C(\mathcal{X}, \mathbb{R}^m) \setminus \mathcal{T}_f\). By definition of \(\mathcal{T}_f\), there exists a pair \((\mathbf{x}^1, \mathbf{x}^2) \in \mathcal{P}_f\) such that
$$
G(\mathbf{x}^1) = G(\mathbf{x}^2).
$$
Since \((\mathbf{x}^1, \mathbf{x}^2) \in \mathcal{P}_f\), we have \(\mathbf{x}^1 \ne \mathbf{x}^2\), and therefore \(\|\mathbf{x}^1 - \mathbf{x}^2\| > 0\). Choose \(n \in \mathbb{N}\) such that
$$
\frac{1}{n} \le \|\mathbf{x}^1 - \mathbf{x}^2\|.
$$
Then \((\mathbf{x}^1, \mathbf{x}^2) \in \mathcal{D}_n\) by definition of \(\mathcal{D}_n\), and since \(G(\mathbf{x}^1) = G(\mathbf{x}^2)\), it follows that \(G \in \mathcal{F}_n\).

Hence every function that fails to satisfy Target Recoverability belongs to at least one of the sets \(\mathcal{F}_n\), which proves the desired inclusion.

\textbf{4. \(\mathcal{T}_f\) is dense in \(C(\mathcal{X}, \mathbb{R}^m)\).}

From 3, we have the inclusion
$$
\mathcal{T}_f \supseteq C(\mathcal{X}, \mathbb{R}^m) \setminus \bigcup_{n=1}^\infty \mathcal{F}_n.
$$

Each set \(\mathcal{F}_n\) is closed and nowhere dense; consequently, the countable union \(\bigcup_{n=1}^\infty \mathcal{F}_n\) is a meager subset of \(C(\mathcal{X}, \mathbb{R}^m)\).  
Since \(C(\mathcal{X}, \mathbb{R}^m)\) equipped with the uniform norm is a complete metric space, it is a Baire space. In any Baire space, the complement of a meager set is dense (in fact, it is a dense \(G_\delta\) set). Therefore,
$$
C(\mathcal{X}, \mathbb{R}^m) \setminus \bigcup_{n=1}^\infty \mathcal{F}_n
$$
is dense in \(C(\mathcal{X}, \mathbb{R}^m)\).

Because \(\mathcal{T}_f\) contains this dense set, it follows that \(\mathcal{T}_f\) itself is dense in \(C(\mathcal{X}, \mathbb{R}^m)\).

In particular, note that \(f \in \mathcal{T}_f\): by definition of \(\mathcal{P}_f\), if \((\mathbf{x}^1, \mathbf{x}^2) \in \mathcal{P}_f\), then \(f(\mathbf{x}^1) \ne f(\mathbf{x}^2)\), so \(f\) trivially satisfies Target Recoverability with respect to itself. Hence, for every \(\varepsilon > 0\),
$$
B_\varepsilon(f) \cap \mathcal{T}_f \ne \emptyset,
$$
which confirms the density of \(\mathcal{T}_f\) in a neighborhood of \(f\).

\end{document}